\documentclass[10pt,twocolumn,letterpaper]{article}

\usepackage{wacv}
\usepackage{times}
\usepackage{epsfig}
\usepackage{graphicx}
\usepackage{amsmath}
\usepackage{amssymb}
\usepackage{booktabs}
\usepackage{enumitem}
\usepackage{subcaption}
\usepackage{longtable}
\usepackage{float}
\usepackage{flushend}
\usepackage{algpseudocode}
\usepackage{algorithm}
\usepackage{placeins}
\usepackage{multirow}
\usepackage{pifont}
\usepackage{bm}
\usepackage{color,soul}
\usepackage[accsupp]{axessibility}  
\let\svthefootnote\thefootnote
\newcommand\freefootnote[1]{%
  \let\thefootnote\relax%
  \footnotetext{#1}%
  \let\thefootnote\svthefootnote%
}

\newcommand{\xmark}{\ding{55}}%

\def\wacvPaperID{1259} 

\wacvapplicationstrack 

\wacvfinalcopy 

\ifwacvfinal
\usepackage[breaklinks=true,bookmarks=false]{hyperref}
\else
\usepackage[pagebackref=true,breaklinks=true,colorlinks,bookmarks=false]{hyperref}
\fi

\begin{document}

\title{\textsc{LineEX}: Data Extraction from Scientific Line Charts}

\author{Shivasankaran V P*, Muhammad Yusuf Hassan*, Mayank Singh\\
IIT Gandhinagar, Gujarat, India\\
{\tt\small vp.shivasan@iitgn.ac.in, \tt\small md.hassan@iitgn.ac.in, \tt\small singh.mayank@iitgn.ac.in } 
~~
}

\maketitle
\thispagestyle{empty}

\begin{abstract}
   In this paper, we introduce \textsc{LineEX} that extracts data from scientific line charts. We adapt existing vision transformers and pose detection methods and showcase significant performance gains over existing SOTA baselines. We also propose a new loss function and present its effectiveness against existing loss functions. In addition, we synthetically created the largest line chart dataset comprising 430K images. The code is available at \url{https://github.com/Shiva-sankaran/LineEX}.
\end{abstract}


\section{Introduction}
\label{sec:intro}
\freefootnote{*Equal contribution} The modern world generates large volumes of data every day, but most of it is unusable due to processing,  representation, and storage challenges. Scientific papers, too, contain a large proportion of non-textual content, such as charts and images, that do not serve more purpose than visualization~\cite{chartocr}. This non-textual content, if successfully processed, can be used in designing high-quality scholarly search engines~\cite{subramanian2020medicat}, machine-generated task-specific leaderboards~\cite{singh2019automated}, and scholarly assistants for impaired people~\cite{sviatov2021data}. With the advent of deep learning architectures and the availability of large volumes of scholarly datasets, we have witnessed a recent surge in efforts to extract data from scientific charts. However, most of these works suffer from prevalent ML-related challenges such as reproducibility and inaccessibility (see Table~\ref{tab:table-survey} for more details). To this end, we aim to develop a chart information extraction system that is entirely reproducible, publicly available and produces high-quality output.

In contrast to existing chart extraction systems that extract information from all possible types of charts, the current work focuses on \textit{line charts}.   We propose a system, hereafter \textsc{LineEX}, that leverages powerful transformer architectures~\cite{vaswani2017attention} for information extraction from line charts. The proposed system is inspired by vision transformers~\cite{vit} that have recently outperformed all existing state-of-the-art systems. \textsc{LineEX} comprises three modules (i) keypoint extraction, (ii) chart element detection and text extraction, and (iii) keypoints grouping, legend mapping and data scaling.  We robustly evaluate each module and compare it with the state-of-the-art ChartOCR system~\cite{chartocr}. In addition to the proposed system, we created the largest synthetic dataset for line chart information extraction. To summarise, the main contributions of our work are: 
\begin{itemize}[noitemsep,nolistsep]
    \item Construction of the largest line chart dataset comprising 430K images incorporating diversity in the number of lines, font size, figure size, placement of legend boxes, line color and marker styles, background, and gridlines. 
    \item Adapting vision transformer architectures for line chart information extraction tasks. 
   \item Showcasing limitations of existing loss functions and proposal of a new loss function that addresses these challenges. 
    \item Legend-to-line mapping by matching line patches with legend markers based on image similarity.
\end{itemize}

\begin{table*}[!tbh]
    \centering
    \resizebox{\hsize}{!}{
    \begin{tabular}{|l|c|c||c|c|c||c|c||c||c||c|c|}
       \hline
       \multirow{2}{*}{System Name}& 
       \multicolumn{2}{c||}{Axes} & 
       \multicolumn{3}{c||}{Lines} & 
       \multicolumn{2}{c||}{Legends} &  
       \multirow{2}{*}{\parbox[t]{1cm}{\centering Chart Title}} & 
       \multirow{2}{*}{\parbox[t]{1.2cm}{\centering Data Scaling}}  &
       \multirow{2}{*}{\parbox[t]{1.8cm}{\centering Legend-line Mapping}} & \multirow{2}{*}{\parbox[t]{1.8cm}{\centering Code-base Availability}}\\
       
       &Title&Ticks&Keypoints&Thickness&Dashing&Marker&Text&&&&\\\hline
       FigureSeer~\cite{figureseer} &\checkmark&\xmark&\checkmark&\checkmark&\checkmark&\checkmark&\checkmark&\xmark&\checkmark&\checkmark&\checkmark\\\hline
       ChartOCR~\cite{chartocr} &\checkmark&\xmark&\checkmark&\checkmark&\checkmark&\xmark&\xmark&\checkmark&\checkmark&\xmark&\checkmark\\\hline
       Linear Programming~\cite{linear} &\checkmark&\checkmark&\checkmark&\xmark&\checkmark&\checkmark&\checkmark&\checkmark&\checkmark&\checkmark&\xmark\\\hline
       LineEX (ours) &\checkmark&\checkmark&\checkmark&\checkmark&\checkmark&\checkmark&\checkmark&\checkmark&\checkmark&\checkmark&\checkmark\\\hline
    \end{tabular}}
    \caption{Comparison of existing chart data extraction methodologies for different line chart components.}\label{tab:table-survey}
\end{table*}

\section{Scientific Line Charts}
\label{sec:line_charts}
\begin{figure}[!tbh]
\centering
  \includegraphics[width=1\linewidth]{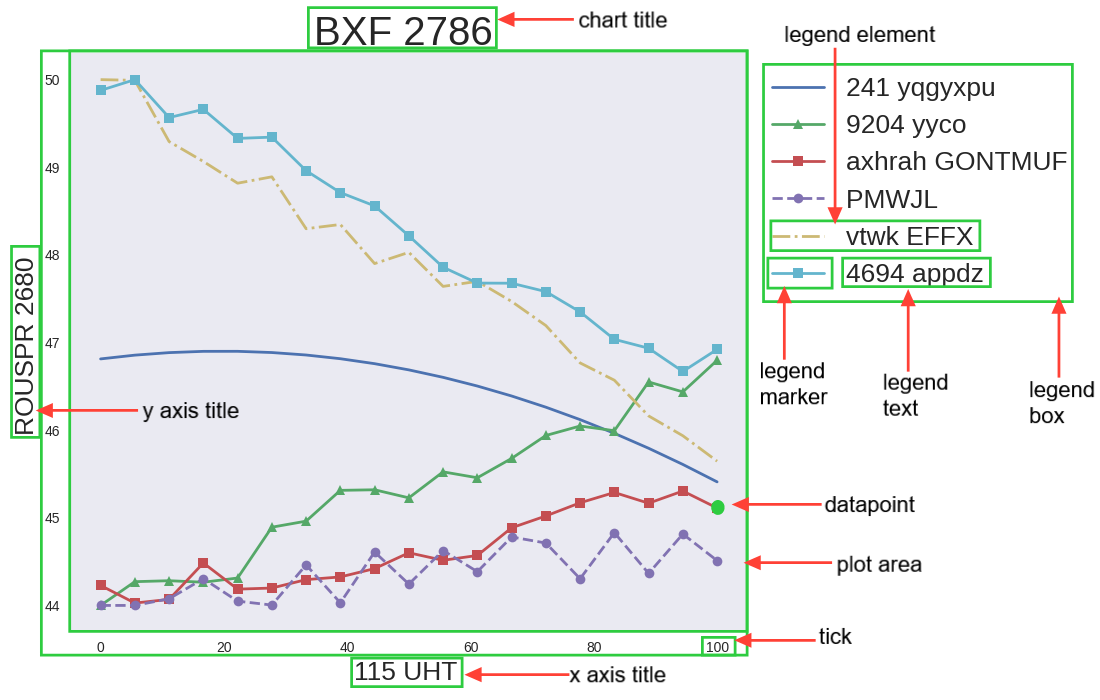}
  \caption{The different components of a line chart.}\label{fig:example_chart}
\end{figure}

Line charts are the most common form of charts found in scientific literature. Line charts are used to visually represent a series of data points or a mathematical function. The main components of a line chart include (i) \textit{axes}, (ii) \textit{lines}, (iii) \textit{a legend box} and (iv) \textit{a chart title}. Optionally, the major components can be further subdivided into sub-components. For example, the legend box comprises visual markers (popularly known as \textit{legend marker}) and associated labels (popularly known as \textit{legend text}), which are required for mapping each line to its corresponding text-marker pair. Similarly, axes comprise titles and ticks. Figure~\ref{fig:example_chart} showcases and describes a simplistic variant of a line chart comprising the above components. 

 For example, charts comprising a legend box inside the chart area with two or more columns for listing text-marker pairs, log scaled axes, or inclined axes ticks. Secondly, we find chart variants comprising different stylistic features, such as lines of different thicknesses, dashed or continuous lines, or lines with or without markers. Lastly, different plotting software can generate different visually-looking charts with the same underlying input. 

Due to high complexities and diversity, the field of chart data extraction has been relatively less explored. Most of the existing chart datasets (see Table~\ref{tab:tablular-datasets} for more details) are synthetic, such as the Adobe Synthetic Chart Dataset~\cite{chart2019} and SYN dataset~\cite{linear}. A few real datasets are curated from the web such as FigureSeer~\cite{figureseer}, ExcelChart400k~\cite{chartocr} and the ICPR-2020~\cite{chart2020}. Real chart datasets are relatively difficult to curate due to problems associated with the manual efforts required to annotate all the components. On the other hand, synthetic chart datasets are generated automatically using plotting software comprising synthetic data values. These datasets are usually more limited in diversity than real datasets but can be constructed in large numbers with minimum human effort.

\begin{table}[b]
 \resizebox{\hsize}{!}{
\begin{tabular}{|l|c|c|c|c|c|}
\hline
 \multirow{2}{*}{\textbf{Name}}&  \multirow{2}{*}{\textbf{Type} }& \multicolumn{2}{|c|}{\textbf{Instances}} &  \multirow{2}{*}{\textbf{Systems}} & \multirow{2}{*}{\textbf{Public}} \\ 

&&Full&Line&&\\ \hline
FigureSeer~\cite{figureseer}  & Real   & 60k        & 60k  &  \cite{figureseer}\cite{chartocr}\cite{linear} &only 1k\\ \hline

Adobe Synthetic~\cite{chart2019} & Synthetic      & 198k & 4.2k& -&\checkmark\\ \hline

ICPR-2020~\cite{chart2020} & Real      & 23k & 10k& -&\checkmark\\ \hline
ExcelChart400k\cite{chartocr}  & Real      & 400k    &          122k&  \cite{chartocr}\cite{linear}  &\checkmark\\ \hline
SYN\cite{linear} & Synthetic & 64k & 64k & \cite{linear} & \xmark \\ \hline
Our dataset  & Synthetic & 430k & 430k &  LineEX &\checkmark \\ \hline
\end{tabular}
}
\caption{Comparison of different chart data extraction datasets.}\label{tab:tablular-datasets}
\end{table}

Given the requirements of extensive manual efforts in curating real datasets, we propose a new line chart dataset. It contains 430k line charts generated from Python's popular \textit{Matplotlib library}\footnote{https://matplotlib.org}. Our dataset comprises charts containing 2 to 6 lines. It also contains annotations for bounding boxes for all possible chart components. Diversity in the generation process includes a random number of lines, font, and figure size. We also implement diversity in the placement of legend boxes. Lines have variations in color and marker style (dots, dashes, and shapes). We generate plots with 27 different plotting styles that bring diversity in line markers, backgrounds, and gridlines. The chart text (chart title, axes labels, legend labels) is randomly generated from alphanumeric strings. Further, we segment our dataset into 400,000 images for training, and 10,000 images for validation and 20,000 images for the test.

\begin{figure*}
\centering
\begin{subfigure}{.4\textwidth}
  \centering
  \includegraphics[width=.9\linewidth]{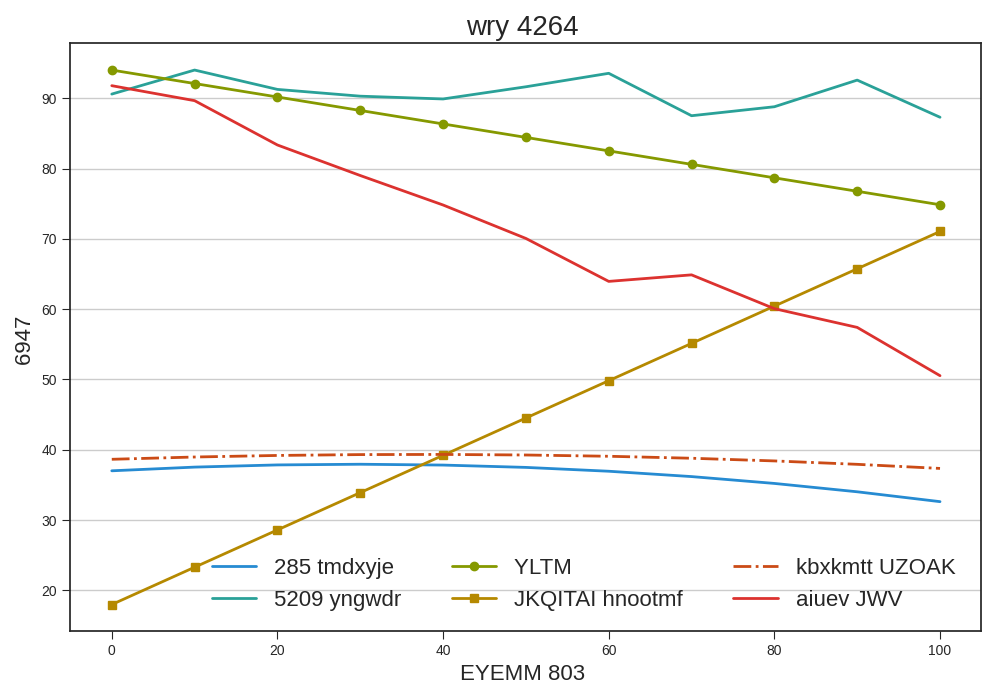}
  \caption{White background with six lines}
  \label{fig:sub1}
\end{subfigure}
\begin{subfigure}{.4\textwidth}
  \centering
  \includegraphics[width=.8\linewidth]{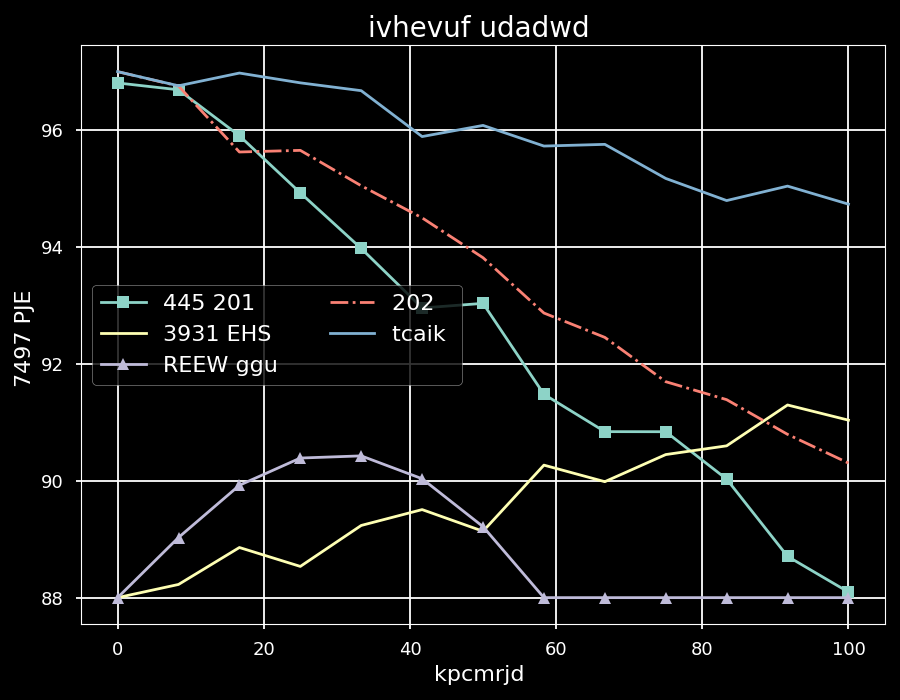}
  \caption{Dark background with five lines}
\end{subfigure}
\caption{Representative line charts from our dataset with different plotting styles. The two charts have different legend placement coordinates.}
\label{fig:test}
\end{figure*}

Table~\ref{tab:tablular-datasets} compares our dataset against existing datasets. It is the largest among existing line chart datasets and approximately 4X larger than ExcelChart400k~\cite{chartocr}. In contrast to the existing synthetic line chart datasets such as Adobe Synthetic Chart Dataset~\cite{chart2019} and SYN dataset~\cite{linear}, our dataset also contains variations in the plotting styles, with charts ranging from darker to brighter backgrounds, bounding boxes around plot area and legend area. Additionally, Adobe Synthetic Chart Dataset does not map the legend marker and text pair to the respective lines. In contrast to the largest existing real dataset ExcelChart400k~\cite{chartocr}, our dataset also contains annotations for legend markers, legend texts, tick texts and their corresponding boxes. Figure~\ref{fig:test} shows two representative line charts from our dataset with different background, tick and legend placement styles.

\section{Related Work}
\label{sec:rel_work}
Most earlier chart data extraction methods used heuristic and rule-based techniques to extract data according to the different chart types. The works by \cite{revision} and \cite{gao} first classify the chart type using a Support Vector Machine, which is then followed by the use of heuristic based data extraction methods. Such rule-based methods have become less prominent due to the lack of generalisability to real-world data and the need to hand-engineer features. To improve results, some hybrid methods like ChartSense \cite{chartsense} also include some human feedback in the data extraction process. These hybrid methods give better results, but they also use human time and labour and thus cannot be implemented into an automated pipeline.

Recently, efforts in deep learning based models have been shown to give better results in chart data extraction. These methods usually build upon object detection models and adapt the processing according to the chart type. ChartText \cite{charttext} uses a Convolutional Neural Network to classify the chart type, then runs object detection models to extract text and labels. Then, type-specific image processing is done to extract the data. ChartOCR \cite{chartocr} also uses a CNN to classify the chart type, which is then followed by a keypoint detection model to extract certain pre-defined keypoints based on the chart type. Postprocessing is done based on the detected chart type to extract chart data using the identified keypoints. These methods perform better than rule-based methods, but they still give problematic results on real-world data. Also, the ChartOCR model does not allow for legend mapping, i.e. identifying which line corresponds to which legend marker or label. Recently, ChartOCR scores are  outperformed by linear programming-based approach~\cite{linear}. However, the source code and trained models are not publicly available.

\section{The LineEX System}
\label{sec:method}

The LineEX system comprises a highly modular pipeline. It consists of three main modules: (i) keypoint extraction, (ii) chart element detection and text extraction, and (iii) keypoints grouping, legend mapping and data scaling.

\begin{figure*}[!tbh]
\centering
  \includegraphics[width=0.7\linewidth]{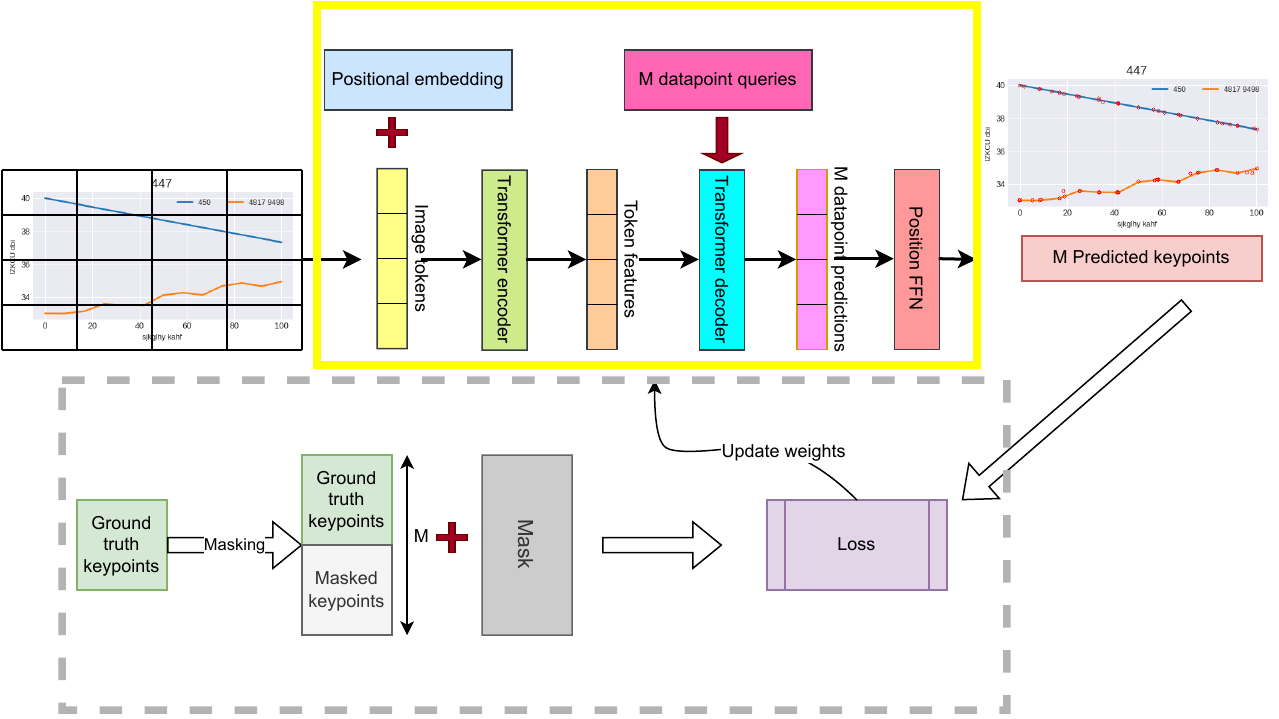}
  \caption{Keypoint detection architecture.}
  \label{fig:keypoint_detection_model}
\end{figure*}

\subsection{Keypoint Extraction} 
\label{sec:keypoint_extraction}
This module extracts keypoints that are joined in order to create lines. For keypoint extraction, we adopt the PE-former~\cite{pe-former} architecture. PE-former is an end-to-end encoder-decoder transformer architecture proposed for human pose estimation. The best performing PE-former variant comprises Cross-Covariance Image Transformer (XCiT)~\cite{xcit} as the encoder and DETR~\cite{detr} based transformer decoder. Figure~\ref{fig:keypoint_detection_model} describes the architecture of the keypoint extraction module.

\vspace*{-0.3cm}
\subsubsection{The Architecture}
\label{sec:architecture}
In contrast to the COCO dataset's 17 joint types, \textsc{LineEX} contains a single class representing a keypoint. Thus, we do not pass the output tokens from the decoder through a feed-forward neural network (FFN) based classification head. However, we leverage FFN to regress each output token to two scalar values (x, y) in the range of [0, 1]. Note that the number of ground truth keypoints in a chart varies drastically based on the number of lines and the complexity of the lines in the chart. A chart with a simple single line typically has 6--10 keypoints, whereas a chart with 4--5 complex lines might contain more than 60 keypoints. We fix the decoder input as M keypoint queries, where M = 64 keypoints. As consistency in the batch size is not possible due to the diversity in the charts, we add multiple dubious keypoints as (0,0) to the ground truth keypoints so that every chart has exactly 64 keypoints.
In some cases, the number of ground truth keypoints is more than 64. In these cases, we only take the first 64 keypoints as our ground truth. In both these cases, we mask them appropriately, with 1 indicating a true ground truth keypoint and 0 indicating a dubious keypoint. This mask is later used during loss calculation so that the model's weights are not affected by the addition of dubious keypoints.

\subsubsection{The Loss Function}
\label{sec:loss_function}
\begin{figure}[b]
\centering
 \includegraphics[width=0.9\linewidth]{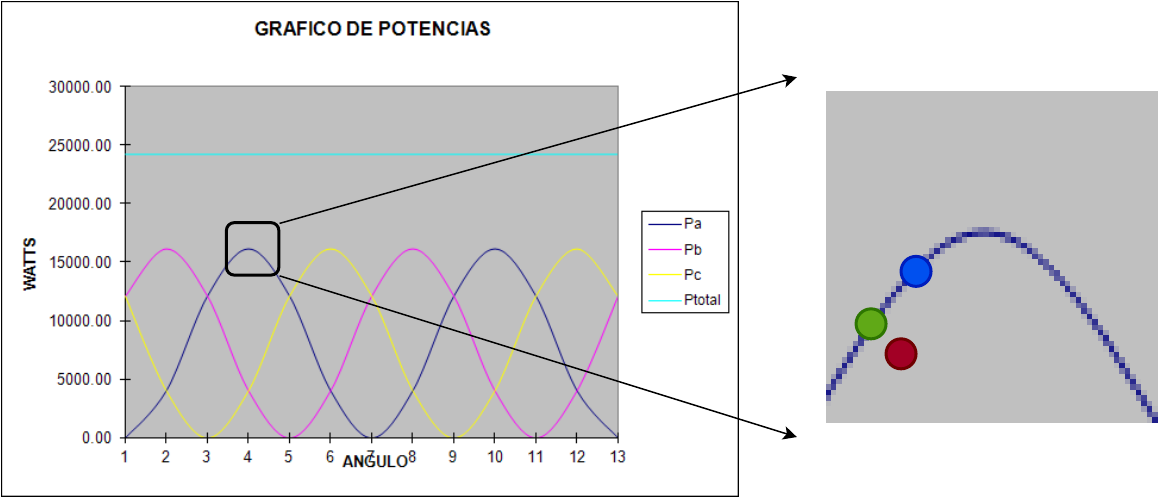}
  \caption{Comparison between predicted keypoints. Ground truth keypoint is denoted by Green color. Predicted keypoints are denoted by Red and Blue colors.} \label{fig:l1_bad}
\end{figure}

Unlike 2D pose estimation problems, the precision of keypoint detection is crucial as the predicted keypoint should lie on the correct line and not in the nearby region. Distance-based loss functions such as L1 or L2 fail to capture this intuition. The problem is illustrated in Figure~\ref{fig:l1_bad}. The predicted keypoint (in Red color) is closer to the ground truth keypoint (in Green color) than the other predicted keypoint (in Blue color). However, as keypoint in Blue color lies on the line, it is an ideal candidate than the keypoint in Red color. To capture this intuition, we propose a new loss function and term it as \textbf{Angular Similarity Error ($\mathcal{L}_{\rm A}$)} function. 

Before formally defining $\mathcal{L}_{\rm A}$, we define \textit{anchor points}. Each keypoint in the training line chart has an associated anchor point. The anchor points are determined through a neighborhood pixel search around the ground truth keypoints. Specifically, in our implementation, we define an anchor point as a point that lies on the same line as the ground truth keypoint and is within 5 pixels away from the ground truth keypoint. All the anchor points are computed and stored in the training data. The estimated first derivative using anchor points is then used to calculate the angular separation between the predicted keypoint and the ground truth keypoint. 

\begin{figure}[!tbh]
\centering
\includegraphics[width=0.4\linewidth]{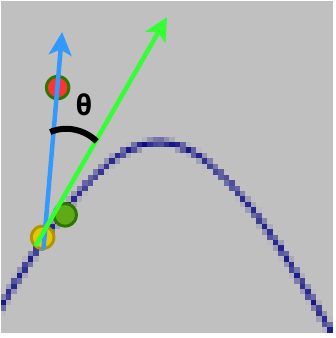}
  \caption{Computing angular similarity error:  Anchor point, ground truth keypoint and predicted keypoint are denoted by Yellow, Green and Red color, respectively. Green arrow is a vector that joins the anchor point and the ground truth keypoint. The Blue arrow is a vector joining the ground truth keypoint and the predicted keypoint. The angle between these vectors is denoted by $\theta$.}
  \label{fig:angle}
\end{figure}

We now define the $\mathcal{L}_{\rm A}$ as the angular deviation of the vector joining the predicted keypoint and the ground truth from the first derivative of the line at the ground truth keypoint. The first derivative at the ground truth keypoint is computed using Newton's difference quotient method. The anchor point is used to evaluate the first derivative of the line at the ground truth keypoint. Figure~\ref{fig:angle} illustrates this idea visually. Consider the ground truth, predicted and anchor point as \textbf{Y}, \boldsymbol{$\hat{Y}$}, and \textbf{A}, respectively. $\mathcal{L}_{\rm A}$ is formally defined as:

\begin{equation}
\mathcal{L}_{\rm A} = 1 - cos(\boldsymbol{Y}-\boldsymbol{A}, \boldsymbol{Y}-\boldsymbol{\hat{Y}})
\end{equation}

We combine $\mathcal{L}_{\rm A}$ with standard L1 loss ($\mathcal{L}_{\rm D}$) to define the overall loss ($\mathcal{L}$) as: 

\begin{equation}
\label{eq:new_loss}
\mathcal{L} = \alpha\mathcal{L}_{\rm D} +  (1-\alpha) \mathcal{L}_{\rm A}
\end{equation}
where, $\alpha$ is a  hyperparameter.

We propose two variants of our keypoint extraction module, one with $\alpha = 1$ (hereafter referred to as $\textsc{LineEX}_D$) and other with $\alpha =0.99$ (hereafter referred to as $\textsc{LineEX}_{D+A}$). $\textsc{LineEX}_D$ is a pure L1 loss-based module, whereas $\textsc{LineEX}_{D+A}$ combines L1 loss with proposed angular similarity error\footnote{We conduct a binary search to identify the optimal $\alpha$ value.}. 

\subsection{Chart Element Detection and Text Extraction} 
\label{sec:bounding_box_Detection}
The second module identifies axes, legend box, chart title, and their sub-components (described in Section~\ref{sec:line_charts}). Specifically, it creates a bounding box around the chart title, axes labels, plot area, axes ticks, and legend markers and text. The extracted information from this module helps in several downstream tasks, such as scaling the extracted keypoints from pixel coordinates to the original coordinates and legend mapping. This module is divided into two sub-modules: 
\begin{itemize}[noitemsep,nolistsep]
    \item \textbf{Detection:} We leverage the original DEtection TRansformer (DETR)~\cite{detr} implementation for predicting bounding boxes around the chart components. DETR is a recently-developed approach that applies the transformer encoder and decoder architecture to object detection and achieves promising performance. To our knowledge, this is the first adoption of the transformer architecture for the problem of chart data extraction. We use the original DETR transformer architecture without making any changes to it, except for adapting its last layer for the following ten classes: (i) Chart title,  (ii) X-axis title, (iii) Y-axis title,  (iv) Ticks, (v) Plot area, (vi) Inner plot area, (vii) Legend box, (viii) Legend marker, (ix) Legend label, and (x) Legend element. We use a pretrained DETR model and finetune it.
    \item \textbf{Text Extraction:} This sub-module generates chart title, axes labels, and legend texts. We pass the input chart through the EasyOCR~\cite{easyocr} tool to read and locate text. EasyOCR outputs the detected text and its corresponding bounding boxes, while the DETR model outputs bounding boxes for each element class. We classify and locate text in the chart by finding the intersection between these bounding boxes. If EasyOCR cannot detect text for a text box detected by DETR, we report that box without any accompanying text.
\end{itemize}

\begin{figure}[!tbh]
\centering
 \includegraphics[width=1\linewidth]{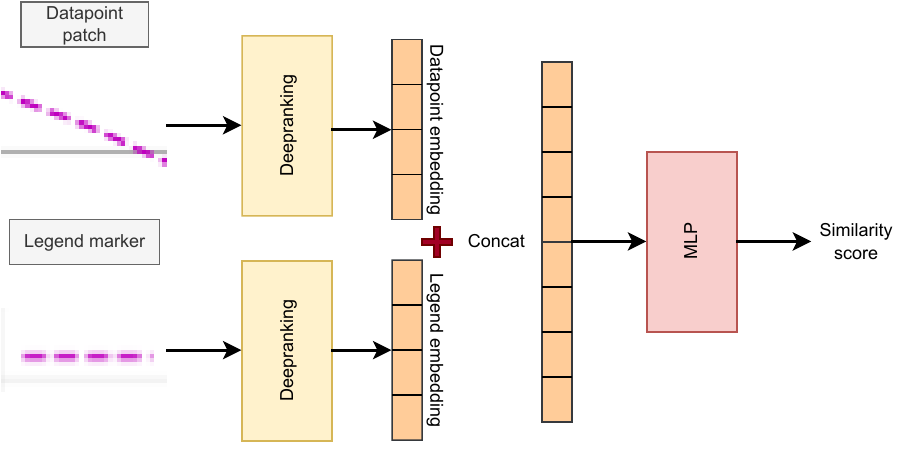}
  \caption{Keypoint grouping and legend mapping module.} \label{fig:leg_map}
\end{figure}

\subsection{Keypoint Grouping, Legend Mapping, and Datapoint Scaling}
\label{sec:grouping_mapping_module}
This module performs three postprocessing tasks, keypoint grouping, legend mapping, and data scaling. We perform keypoint grouping and legend mapping jointly.

\begin{itemize}[noitemsep,nolistsep]
\item \textbf{Keypoint grouping and legend mapping}:  We adapt concepts of image similarity to group keypoints and map them to legend markers. We train the Deepranking model~\cite{deepranking} by constructing triplets of the keypoint patch and its corresponding positive and negative legend markers. The Deepranking model minimizes hinge loss to group keypoint patches that belong to the same legend marker and differentiate them from keypoint patches that belong to different legend markers. We modify the last linear layer's output size from 4096 to 100 to generate a 100-dimensional embedding vector. Next, we train a simple MLP model to predict the similarity score between the embeddings of a keypoint patch and a legend marker. Figure~\ref{fig:leg_map} describes the keypoint grouping and legend mapping module. 

We have the keypoint coordinates and the legend marker bounding boxes during the test phase. We sample 20$\times$40 pixel patches around each detected keypoint and also resize legend marker bounding boxes to 20$\times$40 pixels. Let N be the number of detected legend markers and M be the number of detected keypoints. We send each combination of the keypoint patch and legend marker patch through the above MLP model and obtain a similarity score. This way, we have an N$\times$M matrix of similarity scores between keypoint patches and legend markers. Based on this N$\times$M matrix, we map each keypoint to a unique legend marker. Points at the intersection of two more lines will arbitrarily get mapped to one of the intersecting lines. Every keypoint belonging to the legend marker gets grouped as a detected line, and implicitly, legend mapping is also done.

\begin{algorithm}
\caption{Datapoint Scaling}
\begin{algorithmic}
\State \textbf{Given:} list $T$ (extracted tick values), list $C$  (extracted tick coordinates), list $dps$ (unscaled datapoints)
\State $r\textunderscore list \gets [] $ \Comment{list of ratios corresponding to different ticks pairs}
\For{$i$ in $range(len(C))$}
    \For{$j$ in $range(len(T))$}
        \State $r\textunderscore list.append($abs$((T[j]-T[i])/(C[j]-C[i])))$
    \EndFor
\EndFor
\State Find median ratio ($r_{med}$) and its corresponding indices ($med_{idx}$) from $coords$
\State $scaled$ $\gets  (dps - C[med_{idx}])*r_{med} + T[med_{idx}]$
\end{algorithmic}
\label{algo:datapoint_scaling}
\end{algorithm}

\item \textbf{Datapoint Scaling}: The keypoints extracted in Section~\ref{sec:keypoint_extraction} are present in pixel coordinates. In order to transform the pixel coordinates into their corresponding raw data points, we perform a final step of data scaling, which uses the OCR results and tick coordinates obtained in Section~\ref{sec:bounding_box_Detection}. Algorithm~\ref{algo:datapoint_scaling} takes in the predicted keypoints, the extracted tick values, and their texts and outputs the scaled datapoint values. Note that X-axis ticks and Y-axis ticks need to be run separately through the algorithm as they may have different scaling.
\end{itemize}

\section{Experiments}
\label{sec:exp}
\subsection{Baselines}
\label{sec:baseline}
We list existing systems in Table~\ref{tab:table-survey}. We only experiment with ChartOCR~\cite{chartocr}. ChartOCR is a state-of-the-art system that combines a deep framework and rule-based methods for chart identification and data extraction. We use the trained model available at the project's Github repository\footnote{\url{https://github.com/soap117/DeepRule}}. The other closer work, titled \textit{Linear Programming}~\cite{linear} can not be compared due to the unavailability of source code\footnote{We requested authors, but due to confidentiality clause, the model and the source code cannot be public.}.

\subsection{Datasets}
\label{sec:datasets_eval}
We list chart information extraction datasets in Table~\ref{tab:tablular-datasets}. Due to the unavailability or incomplete ground truth information associated with several datasets, we only conducted experiments on four datasets. 
\begin{enumerate}[noitemsep,nolistsep]
    \item \textbf{Adobe Synthetic}~\cite{chart2019}: The Adobe Synthetic lines test set is used in evaluating the keypoint extraction and the chart element detection modules. The test split comprises of 200 instances. 
    \item \textbf{ExcelChart400k}~\cite{chartocr}: The ExcelChart400k train set was used for training and evaluating the keypoint detection module. The train, validation and test split comprises of 116745, 3074 and 3072 instances, respectively.
    \item \textbf{FigureSeer}~\cite{figureseer}: The FigureSeer test set was used in evaluating the chart element detection module. The test set comprises of 1000 instances. 
    \item \textbf{Our Dataset}: Our synthetic dataset was used in training and evaluating the keypoint extraction and the chart element detection modules. The train, validation and test split comprises of 400000, 10000 and 20000 instances, respectively.
\end{enumerate}

\subsection{Evaluation Metrics}
\label{sec:eval_metrics}

\noindent \textbf{Keypoint Extraction}: 
We evaluate the keypoint extraction module by matching predicted keypoints against the ground truth keypoints. Here, we experiment with two variants of similarity metrics both, based on Object Keypoint Similarity score (OKS)~\cite{OKS}. Let $P = [p_1, p_2, p_3, ..., p_M]$ be $M$ predicted keypoints and let $G = [g_1, g_2, g_3, ..., g_m]$ be $m$  ground truth keypoints. For each predicted keypoint $p_i$, we find the closest ground truth keypoint $g_j$ and the corresponding euclidean distance $d_i$. Let the length of the chart diagonal be $s$. Then, OKS for the predicted keypoint $p_i$ is formally defined as:
    \begin{equation}
OKS(p_i) = \exp{(-\frac{{d_i}^2}{2{s}^2{k}^2})}
\end{equation}
where k is a breathing zone for the predicted point to be classified as ground truth keypoint~\cite{OKS}. We can assign k such values that can enforce stricter bounds. The first similarity metric variant is stricter than the second one. We describe the two variants below: 

\begin{itemize}[noitemsep,nolistsep]
    \item $\mathbf {sim_{str}}$: As the total number of predicted keypoints is significantly higher than the number of ground truth keypoints; we only match one predicted keypoint with one ground truth keypoint. For strict bounds, we use $k=0.025$~\cite{OKS}. We define $\gamma$ as the threshold to compare the OKS value of a predicted keypoint. We set $\gamma = 0.5$. We assign a true positive label to a $p_i$ if $OKS(p_i) >\gamma$ and the closest ground keypoint is not already assigned to some other predicted keypoint. The corresponding ground keypoint is marked as assigned if the assignment is successful. 
    \item $\mathbf {sim_{rel}}$: The predicted keypoint is given a second chance to be labelled as a true positive in the relaxed version. Here, we consider ground truth line segments before assigning a false positive\footnote{We consider all predicted keypoints as positive. Their assignment to ground truth keypoints can be treated as True or False.} label to the predicted keypoint. Let $g_k$ be the next point in the line in which $g_j$ lies. We define an additional threshold $\beta$. A predicted keypoint is assigned as a true positive label after failing the first OKS threshold if the perpendicular distance from $p_i$ to the line segment joining $g_j$ and $g_k$ is less than $\beta$. Keeping $\beta = 0.007 \times s$ in the current settings gave the best results. 
\end{itemize}

We calculate recall, precision, and F1 score for each image in the dataset using both the strict and relaxed versions. The final metric scores are averaged over all the images in the test dataset.

One major drawback of our model variants is predicting 64 keypoints irrespective of the number of lines present in the chart image. This can lead to low precision scores. In order to improve precision, we deploy a background detection algorithm. The algorithm detects if the predicted keypoint is on or close to any line in the chart. Let $p_i$ be a predicted keypoint, $h$ and $w$ be the height and width of the image. We take a square patch $P$ centred on $p_i$ with a length of $\max(h,w)\times0.05$. We further divide $P$ into nine equal smaller square sections. We compare image histograms of the nine sections of the patch using OpenCV's \textit{compareHist} function~\cite{Opencv_hist} with the co-relation method. If the image histograms of all nine sections are extremely similar, the predicted keypoint is likely not on or close to a line. Algorithm~\ref{algo:BG_detection} describes the background detection in detail.

\begin{algorithm}[t]
\caption{Background Detection}
\begin{algorithmic}
\State $SecHists$ = CalculateColourHistograms($sections$)
\State $threshold$ = 0.98

\For{$hist_1$ in $SecHists$}
    \For{$hist_2$ in $SecHists$} 
        \State $val$ $\gets$ CompareHist($hist_1$,$hist_2$)
        \If{$val < threshold$} 
            \State \Return $False$

        \EndIf 
    \EndFor
\EndFor

\State \Return $True$
\end{algorithmic}
\label{algo:BG_detection}
\end{algorithm}

\begin{table*}[!tbh]
\centering
\resizebox{\textwidth}{!}{
\begin{tabular}{lll|l|l|l|l|l|l|l|l||l|l|l}
&\multirow{2}{*}{} &  \multicolumn{3}{c}{\textbf{ExcelChart400K}} & \multicolumn{3}{c}{\textbf{Adobe Synthetic}} & \multicolumn{3}{c}{\textbf{Ours}} & \multicolumn{3}{c}{\textbf{Smooth}} \\ 
&\textbf{} & Recall & Prec & F1  & Recall & Prec & F1  & Recall & Prec & F1 & Recall & Prec & F1 \\ \hline
\multirow{3}{*}{$sim_{str}$}&ChartOCR           & \textbf{0.85} & \textbf{0.98} & \textbf{0.90}          & 0.76 & \textbf{0.72} & 0.71          & 0.71 & \textbf{0.90} & 0.78 & 0.36 & \textbf{0.74} & 0.46           \\ 
&$\textsc{LineEX}_{D}$   & 0.82 & 0.69 & 0.70          & 0.91 & 0.54 & 0.64          & 0.83 & 0.75  & 0.76 & 0.70 & 0.49 & 0.56          \\ 
&$\textsc{LineEX}_{D+A}$ & 0.84 & 0.80 & 0.78          & \textbf{0.94} & 0.67 & \textbf{0.74}          & \textbf{0.86} & 0.84 & \textbf{0.83} & \textbf{0.72} & 0.52 & \textbf{0.59}            \\ \hline\hline
\multirow{3}{*}{$sim_{rel}$}&ChartOCR           & \textbf{0.85} & \textbf{0.98} & \textbf{0.90}          & 0.78 & 0.80 & 0.76          & 0.74 & \textbf{0.97} & 0.83 & 0.38 & \textbf{0.78} & 0.49          \\ 
&$\textsc{LineEX}_{D}$     & 0.83 & 0.87 & 0.83          & 0.93 & 0.76 & 0.81          & 0.85 & 0.92  & 0.87  & 0.75 & 0.58 & 0.64         \\
&$\textsc{LineEX}_{D+A}$  & 0.85 & 0.90 & 0.85          & \textbf{0.93} & \textbf{0.81} & \textbf{0.84}          & \textbf{0.87} & 0.94 & \textbf{0.89}   & \textbf{0.77} & 0.61 & \textbf{0.67}         \\ \hline
\end{tabular}}
\caption{Keypoint extraction performance comparison between our proposed method and ChartOCR.}\label{tab:KP_eval}
\end{table*}

\begin{table*}[!tbh]
\resizebox{\textwidth}{!}{
\begin{tabular}{ll|c|c|c|c|c|c|c|c|c|c}

\multirow{2}{*}{Dataset}     & \multirow{2}{*}{Model} & \multirow{2}{*}{\parbox[t]{1cm}{\centering Legend Box}}     & \multirow{2}{*}{\parbox[t]{1cm}{\centering Y-axis Title}}   & \multirow{2}{*}{\parbox[t]{1cm}{\centering Chart Title}}    & \multirow{2}{*}{\parbox[t]{1cm}{\centering X-axis Title}}   & \multirow{2}{*}{\parbox[t]{1cm}{\centering Plot Area} }     & \multirow{2}{*}{\parbox[t]{1.5cm}{\centering Inner Plot Area}} & \multirow{2}{*}{Ticks}          & 
\multirow{2}{*}{\parbox[t]{1cm}{\centering Legend Marker}}  & \multirow{2}{*}{\parbox[t]{1cm}{\centering Legend Label}}   & \multirow{2}{*}{\parbox[t]{1.5cm}{\centering Legend Element}} \\ 
&&&&&&&&&&\\\hline
\multirow{2}{*}{Our dataset} & ChartOCR& 89.07& \textbf{99.52} & 96.03& 85.81& 99.69& 97.85& - & - & -& - \\ 
 & \textsc{LineEX}              & \textbf{99.97}& 83.84          & \textbf{100.0} & \textbf{100.0} & \textbf{100.0} & \textbf{99.97}  & \textbf{85.55} & \textbf{82.39} & \textbf{82.71} & \textbf{83.74}   \\ \hline
\multirow{2}{*}{Adobe Synthetic} & ChartOCR               & -              & 100.0         & 81.00          & 59.50          &-             & 100.0           & -              & -              & -              & -             \\ 
 & \textsc{LineEX}             & -              & 100.0          & \textbf{99.6}  & \textbf{100.0} & -              & 100.0  & \textbf{57.52} & \textbf{74.04} & \textbf{88.94} & -             \\ \hline
\multirow{2}{*}{FigureSeer}  & ChartOCR               & -              & \textbf{97.66}          & -              & 80.96         & -             & \textbf{99.13}           & -             & -              & -              & -             \\ 
       & \textsc{LineEX}               &-              & 70.79         & -              & \textbf{96.28}          & -              & 96.38          & -              & \textbf{54.26}          & \textbf{63.96}          & -             \\ \hline
\end{tabular}}
\caption{Comparison of mAP scores between our proposed model and ChartOCR. As not all models detect all chart components and not each dataset contain all of these chart components, we denote those cells by `-'.}\label{tab:element_detection_results}
\end{table*}

\noindent \textbf{Chart Element Detection}
Here, we measure the performance of our second module (see Section~\ref{sec:bounding_box_Detection} for more details). We use the standard mean Average Precision (mAP) metric for measuring and comparing the performance of different systems. The mAP score is defined as the mean of the Average Precision of each object class in the dataset. The average precision of a class is the weighted mean of the precision values for detecting a bounding box at each confidence threshold. We use the public implementation~\cite{mAP} to evaluate the chart element detection module.

\noindent \textbf{Keypoint Grouping}
Here, we evaluate our third module (see Section~\ref{sec:grouping_mapping_module} for more details) which groups detected keypoints into the lines. We use the modified definition of F1 score~\cite{chartocr} to measure keypoint grouping. 

\noindent \textbf{Legend Mapping}
Finally, we evaluate the last module of the pipeline (see Section~\ref{sec:grouping_mapping_module} for more details) that maps each detected legend marker to a line that was grouped in the previous step. We use the standard F1 score weighted by the number of lines in each chart.

\section{Results and Discussions}
\label{sec:res_disc}

\begin{figure*}[!tbh]
\centering
 \includegraphics[trim=90 190 0 50, clip,width=0.85\linewidth]{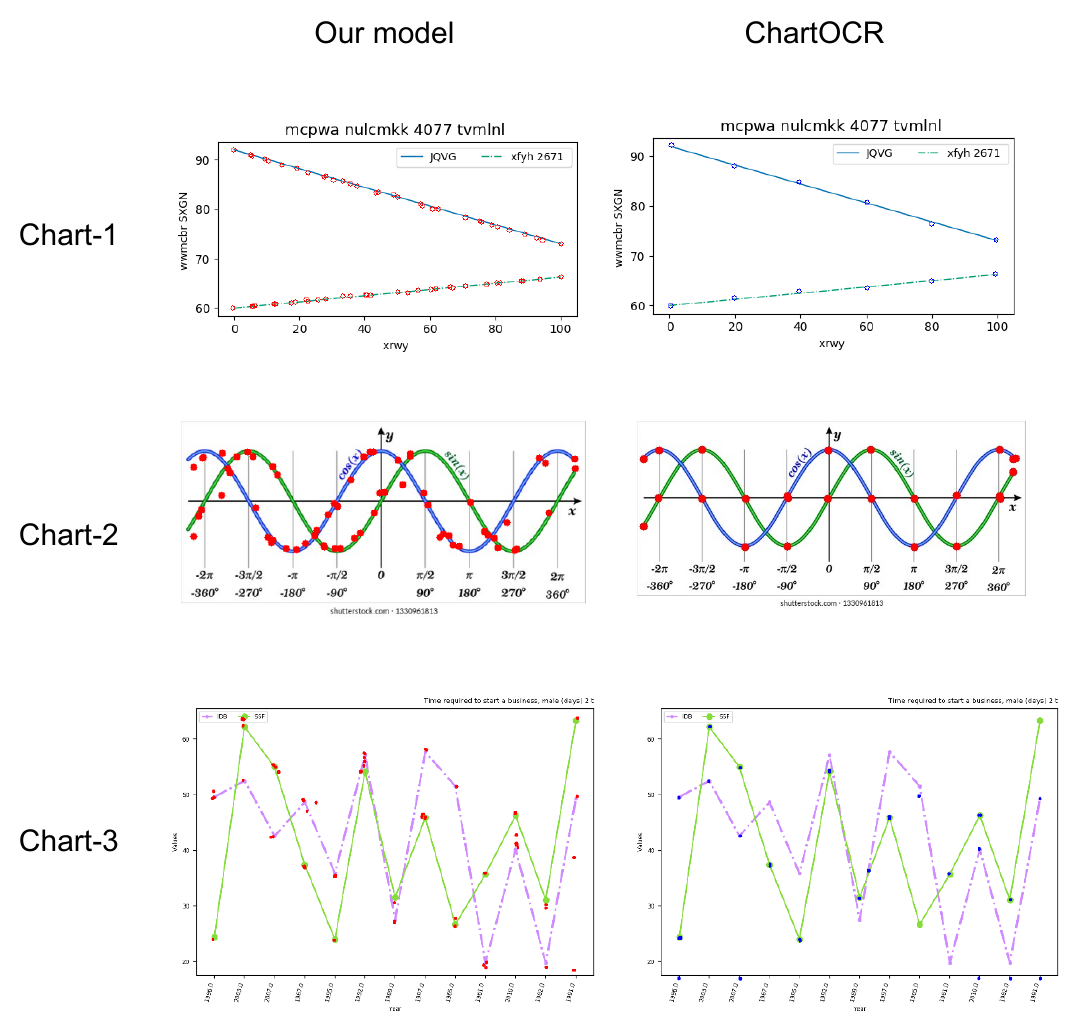}
  \caption{Column 1: \textsc{LineEX} output, Column 2: ChartOCR output.} \label{fig:kp_comp_charts}
\end{figure*}

\noindent \textbf{Keypoint Extraction}
\label{sec:KPE_results}
Table~\ref{tab:KP_eval} presents performance scores for keypoint extraction. Among the three systems, $\textsc{LineEX}_{D+A}$ performed best. ChartOCR outperforms \textsc{LineEX} on ExcelChart400k dataset. However, it performs poorly on other datasets. We attribute this behaviour to its poor generalization ability (ChartOCR is trained using ExcelChart400k dataset). Simple line charts, such as the charts in the first row in Figure~\ref{fig:kp_comp_charts}, having less number of ground truth keypoints, results in lower precision for \textsc{LineEX}. However, complex line charts such as sine or cosine graphs (see second row in Figure~\ref{fig:kp_comp_charts}), containing a large number of keypoints yields high precision scores for \textsc{LineEX}. Empirically, we also find that ChartOCR performs poorly on smooth lines. We validate this finding by conducting an experiment on a 3000 chart images dataset~\cite{APEX}. Each chart comprises multiple smooth lines formed using B-spline and cubic spline interpolated curves. The last three columns in Table~\ref{tab:KP_eval} validate our empirical finding.

\textbf{Chart Element Detection}
\label{sec:element_results}
Table~\ref{tab:element_detection_results} compares performance scores for chart element detection. $\textsc{LineEX}$ performs better than ChartOCR on the Adobe Synthetic dataset and our dataset. Both systems perform comparably on the FigureSeer dataset. In contrast to ChartOCR, $\textsc{LineEX}$ supports the detection of ticks and legend mapping elements (markers and corresponding text). We note that it is difficult to achieve a high mAP score for elements like ticks, legend markers, and legend text due to numerous of them being present in a single chart.

\textbf{Keypoint Grouping}
\label{sec:group_results}
Table~\ref{tab:grouping_results} compares the keypoint grouping performance. Both systems showcase similar performance. One of the reasons for poor performance on the Adobe dataset is the existence of less popular symbols like cross marks, stars, plus signs, and vertical lines in legend markers. We do not evaluate on ExcelChart400K dataset due to the unavailability of legend marker annotations. 

\textbf{Legend Mapping}
\label{sec:map_results}
{The F1 score for mapping predicted lines to legend markers was obtained as 0.79.}

\textbf{Ablation Experiments} We conduct two ablation experiments. We replace the keypoint extraction transformer encoder in the first experiment with a Resnet50 encoder~\cite{Renset}. In contrast to the Resnet50 encoder, the transformer encoder based $\textsc{LineEX}$ variant performs 4X better (see Supplementary). In the second experiment, we replace the grouped predicted keypoints with the grouped ground-truth keypoints for legend mapping. We achieve an F1 score of 0.87. This shows that errors introduced in keypoint extraction and grouping stage can lead to low mapping scores.

\begin{table}[!tbh]
\centering
{%
\begin{tabular}{l|c|c}

& Adobe Synthetic  & Our Dataset \\ \hline
ChartOCR & 0.54 & 0.93 \\ 
$\textsc{LineEX}_{D+A}$ & 0.54 & 0.93 \\ \hline
\end{tabular}
}
\caption{F1 scores for keypoint grouping task.}\label{tab:grouping_results}
\end{table}

\section{Conclusion}
\label{sec:con}
In this paper, we proposed \textsc{LineEX} for extracting data from scientific line charts using vision transformers. 
In the future, the proposed work can be expanded into two possible directions: (i) extending to other chart types and (ii) proposal of downstream usecases. The downstream usecases require more precise extraction models, which could be an interesting research direction to explore.
{\small
\bibliographystyle{ieee_fullname}
\bibliography{egbib}

@article{figureseer,
  author		= "Noah Siegel and Zachary Horvitz and Roie Levin and  Santosh Divvala and Ali Farhadi ",
  title			= "FigureSeer: Parsing Result-Figures in Research Papers",
  journal		= "ECCV",
  pages			= "664--680",
  year			= "2016"
}

@misc{chart2019,
  author		= "Kenny Davila",
  title			= "Competition on HArvesting Raw Tables (CHART) 2019 - PubMedCentral", 
  year			= "2019",
  note			= "ICDAR-CHART-2019-PMC \url{https://tc11.cvc.uab.es/datasets/ICDAR-CHART-2019-PMC_1}"
}

@misc{APEX,
      title={APEX-Net: Automatic Plot Extractor Network}, 
      author={Aalok Gangopadhyay and Prajwal Singh and Shanmuganathan Raman},
      year={2021},
      eprint={2101.06217},
      archivePrefix={arXiv},
      primaryClass={cs.CV}
 }

@misc{chart2020,
  author		= "Kenny Davila",
  title			= "ICPR 2020 Competition on HArvesting Raw Tables", 
  year			= "2020",
  note			= "ICPR2020-CHART-Info \url{https://tc11.cvc.uab.es/datasets/ICPR2020-CHART-Info_1}"
}

@inproceedings{revision,
author = {Savva, Manolis and Kong, Nicholas and Chhajta, Arti and Fei-Fei, Li and Agrawala, Maneesh and Heer, Jeffrey},
title = {ReVision: Automated Classification, Analysis and Redesign of Chart Images},
year = {2011},
booktitle = {Proceedings of the 24th Annual ACM Symposium on User Interface Software and Technology},
pages = {393--402},
}

@inproceedings{chartsense,
  title = {ChartSense: Interactive Data Extraction from Chart Images},
author={Jung, Daekyoung and Kim, Wonjae and Song, Hyunjoo and Hwang, Jeong-in and Lee, Bongshin and Kim, Bohyoung and Seo, Jinwook},
  booktitle={Proceedings of the 2017 CHI Conference on Human Factors in Computing Systems},
  year = {2017},
pages={6706--6717}
}

@inproceedings{gao,
  title={View: Visual Information Extraction Widget for improving chart images accessibility},
  author={Jinglun Gao and Yin Zhou and Kenneth E. Barner},
  booktitle={19th IEEE International Conference on Image Processing},
  year={2012},
  pages={2865-2868}
}

@inproceedings{vit,
    title={An Image is Worth 16x16 Words: Transformers for Image Recognition at Scale},
    author={Alexey Dosovitskiy and Lucas Beyer and Alexander Kolesnikov and Dirk Weissenborn and Xiaohua Zhai and Thomas Unterthiner and Mostafa Dehghani and Matthias Minderer and Georg Heigold and Sylvain Gelly and Jakob Uszkoreit and Neil Houlsby},
    booktitle={International Conference on Learning Representations},
    year={2021},
    url={https://openreview.net/forum?id=YicbFdNTTy}
}

@Article{mAP,
AUTHOR = {Padilla, Rafael and Passos, Wesley L. and Dias, Thadeu L. B. and Netto, Sergio L. and da Silva, Eduardo A. B.},
TITLE = {A Comparative Analysis of Object Detection Metrics with a Companion Open-Source Toolkit},
JOURNAL = {Electronics},
VOLUME = {10},
YEAR = {2021},
NUMBER = {3},
ARTICLE-NUMBER = {279},
URL = {https://www.mdpi.com/2079-9292/10/3/279},
ISSN = {2079-9292},
DOI = {10.3390/electronics10030279}
}

@article{charttext,
  title={Chart-Text: A Fully Automated Chart Image Descriptor},
  author={Abhijit Balaji and Thuvaarakkesh Ramanathan and Venkateshwarlu Sonathi},
  year={2018},
  journal={arXiv preprint arXiv:1812.10636},

}

@inproceedings{chartocr,
author = {Luo, Junyu and Li, Zekun and Wang, Jinpeng and Lin, Chin-Yew},
title = {ChartOCR: Data Extraction from Charts Images via a Deep Hybrid Framework},
organization = {The Computer Vision Foundation},
booktitle = {2021 IEEE Winter Conference on Applications of Computer Vision (WACV)},
year = {2021},
month = {January},
url = {https://www.microsoft.com/en-us/research/publication/chartocr-data-extraction-from-charts-images-via-a-deep-hybrid-framework/},
}

@inproceedings{sviatov2021data,
  title={Data Extraction of Charts with Hybrid Deep Learning Model},
  author={Sviatov, Kirill and Yarushkina, Nadezhda and Sukhov, Sergey},
  booktitle={International Conference on Computational Science and Its Applications},
  pages={382--393},
  year={2021},
  organization={Springer}
}

@article{vaswani2017attention,
  title={Attention is all you need},
  author={Vaswani, Ashish and Shazeer, Noam and Parmar, Niki and Uszkoreit, Jakob and Jones, Llion and Gomez, Aidan N and Kaiser, {\L}ukasz and Polosukhin, Illia},
  journal={Advances in neural information processing systems},
  volume={30},
  year={2017}
}

@inproceedings{subramanian2020medicat,
  title={MedICaT: A Dataset of Medical Images, Captions, and Textual References},
  author={Subramanian, Sanjay and Wang, Lucy Lu and Bogin, Ben and Mehta, Sachin and van Zuylen, Madeleine and Parasa, Sravanthi and Singh, Sameer and Gardner, Matt and Hajishirzi, Hannaneh},
  booktitle={Findings of the Association for Computational Linguistics: EMNLP 2020},
  pages={2112--2120},
  year={2020}
}

@inproceedings{singh2019automated,
  title={Automated early leaderboard generation from comparative tables},
  author={Singh, Mayank and Sarkar, Rajdeep and Vyas, Atharva and Goyal, Pawan and Mukherjee, Animesh and Chakrabarti, Soumen},
  booktitle={European Conference on Information Retrieval},
  pages={244--257},
  year={2019},
  organization={Springer}
}

@INPROCEEDINGS{linear,

  author={Kato, Hajime and Nakazawa, Mitsuru and Yang, Hsuan-Kung and Chen, Mark and Stenger, Bj{\"o}rn},

  booktitle={2022 IEEE/CVF Winter Conference on Applications of Computer Vision (WACV)}, 

  title={Parsing Line Chart Images Using Linear Programming}, 

  year={2022},

  volume={},

  number={},

  pages={2553-2562},

  doi={10.1109/WACV51458.2022.00261}}

@inproceedings{detr,
  title={End-to-end object detection with transformers},
  author={Carion, Nicolas and Massa, Francisco and Synnaeve, Gabriel and Usunier, Nicolas and Kirillov, Alexander and Zagoruyko, Sergey},
  booktitle={European conference on computer vision},
  pages={213--229},
  year={2020}
}

@article{xcit,
  title={Xcit: Cross-covariance image transformers},
  author={Ali, Alaaeldin and Touvron, Hugo and Caron, Mathilde and Bojanowski, Piotr and Douze, Matthijs and Joulin, Armand and Laptev, Ivan and Neverova, Natalia and Synnaeve, Gabriel and Verbeek, Jakob and others},
  journal={Advances in neural information processing systems},
  volume={34},
  year={2021}
}

@article{pe-former,
  title={PE-former: Pose Estimation Transformer},
  author={Panteleris, Paschalis and Argyros, Antonis},
  journal={arXiv preprint arXiv:2112.04981},
  year={2021}
}

@misc{easyocr,
  author		= "JaidedAI",
  title			= "EasyOCR", 
  year			= "2022",
  note			= "version 1.4.2
  \url{https://github.com/JaidedAI/EasyOCR}"
}

@inproceedings{deepranking,
  title={Learning fine-grained image similarity with deep ranking},
  author={Wang, Jiang and Song, Yang and Leung, Thomas and Rosenberg, Chuck and Wang, Jingbin and Philbin, James and Chen, Bo and Wu, Ying},
  booktitle={Proceedings of the IEEE conference on computer vision and pattern recognition},
  pages={1386--1393},
  year={2014}
}

@inproceedings{Renset,
  title={Deep residual learning for image recognition},
  author={He, Kaiming and Zhang, Xiangyu and Ren, Shaoqing and Sun, Jian},
  booktitle={Proceedings of the IEEE conference on computer vision and pattern recognition},
  pages={770--778},
  year={2016}
}

@misc{Opencv_hist,
  key = {OpenCV},
  title = {OpenCV method for Image histogram Comparison},
  howpublished = {\url{https://docs.opencv.org/3.4/d8/dc8/tutorial_histogram_comparison.html}},
  note = {Accessed: 2022-07-04}
}

@misc{OKS,
  key = {COCO},
  title = {Object keypoint similarity},
  howpublished = {\url{https://cocodataset.org/#keypoints-eval}},
  note = {Accessed: 2022-07-13}
}
}
\flushend
\end{document}